\documentclass[conference]{IEEEtran}
\usepackage{url}
\usepackage{subfigure}
\usepackage{pgfplots}
\usepackage{soul}
\usepackage{color}
\usepackage{cite}
\usepackage{amsmath,amssymb,amsfonts}
\usepackage{algorithmic}
\usepackage{graphicx}
\usepackage{textcomp}
\usepackage{xcolor}
\def\BibTeX{{\rm B\kern-.05em{\sc i\kern-.025em b}\kern-.08em
    T\kern-.1667em\lower.7ex\hbox{E}\kern-.125emX}}
\usepackage{hyperref}
\hypersetup{colorlinks,linkcolor={blue},citecolor={blue},urlcolor={blue}}

\IEEEoverridecommandlockouts

\begin{document}

\title{Learning-Based Video Game Development in MLP@UoM: An Overview\textsuperscript{*}\thanks{\textsuperscript{*} Invited paper presented as a keynote speech in ICEEIE'19.}
}
\author{Ke Chen,~\IEEEmembership{Senior~Member,~IEEE}\\
Department of Computer Science, The University of Manchester, Manchester M13 9PL, U.K.\\
Email: Ke.Chen@manchester.ac.uk}

\maketitle

\begin{abstract}
In general, video games not only prevail in entertainment but also have become an alternative methodology for knowledge learning, skill acquisition and assistance for medical treatment as well as health care in education, vocational/military training and medicine. On the other hand, video games also provide an ideal test bed for AI researches. To a large extent, however, video game development is still a laborious yet costly  process, and there are many technical challenges ranging from game generation to intelligent agent creation. Unlike traditional methodologies, in \emph{Machine Learning and Perception Lab at the University of Manchester} (MLP@UoM), we advocate applying machine learning to different tasks in video game development to address several challenges systematically. In this paper, we overview the main progress made in MLP@UoM recently and have an outlook on the future research directions in learning-based video game development arising from our works.
\end{abstract}


\section{Introduction}

The video games industry has drastically grown since its inception and even surpassed the size of the film industry in 2004. Nowadays, the global revenue of the video industry still rises and increases, and the wide-spread availability of high-end graphics hardware have resulted in a demand for more complex video games. This in turn has increased the complexity of game development. In general, video games not only prevail in entertainment but also have become an alternative methodology for
knowledge learning, skill acquisition and assistance for medical
treatment as well as health care in education, vocational/military
training and medicine. From an academic perspective, video games also provide an ideal test bed, which allows for researching into automatic video game development and testing new AI algorithms in such a complex yet well-structured environment with ground-truth.

Machine learning provides an enabling yet underpinning technology for intelligent system development and has made substantial progresses in last decade. Also, machine learning has been explored in video game development \cite{b1,b2} but many issues in video game development have not yet been investigated systematically with machine learning techniques that could offer alternative yet more effective solutions.

The MLP@UoM lab was established in 2003 when the author joined the University of Manchester. The primary goal and research focus in MLP@UoM lie in developing novel machine learning methods to tackle the fundamental/generic yet challenging issues in machine perception including computer vision and speech information processing. In last five years, one of our main research focuses is learning-based video game development. By means of the state-of-the-art machine learning techniques, we have systematically investigated several challenging problems related to video games, including \emph{procedural content generation}, \emph{serious education game development}, \emph{fast skill measurement}, and \emph{autonomous learnable agents}. In this paper, we are going to overview those progresses made in our study. To facilitate the presentation, this paper self-contains a brief introduction to each of those challenging problems before presenting our own solutions.

The rest of this paper is organised as follows. Sections 2-5 overview our progresses in learning-based procedural content generation, learning-based serious education games, fast skill capture via learning and object-based learnable agents, respectively. The last section describes an outlook on several important open research  problems arising from our research.

\section{Learning-based Procedural Content Generation}
\label{sect:lbpcg}

Procedural content generation (PCG) is a process of generating
content for a video game automatically using algorithms. PCG not only has the potential
to provide a basis built upon by developers but also can provide
an endless stream of content for a player to extend the lifetime of
the game. If used properly, it can reduce the amount of resources
required in video game development. In recent years, a variety of approaches
are developed based on the search-based PCG (SBPCG) framework \cite{b3}. The main idea behind SBPCG is utilising evolutionary computation or other meta-heuristic search algorithms to explore a pre-selected content space towards finding out game content appealing to players. It is well known that the same game content can elicit various
emotional and cognitive experience for players of different
types. Thus, experience-driven PCG is demanded for generating personalized content that optimises players' experience \cite{b4}.
Although substantial progress has been made recently, the existing PCG techniques still encounter several challenging problems \cite{b3,b4}. In particular, SBPCG methods often have to handcraft rules and constraints for search, which is laborious and costly. Thus, most of PCG methods has to work on a confined content space, which could exclude quality yet appealing games outside the pre-selected content space. More recently, machine learning techniques have been applied to PCG \cite{b5}. However, almost all of methods under the PCGML framework cannot generate video games from a scratch since those methods have to use the quality handcrafted game content to train learning models for PCG.

\subsection{Learning-based PCG Framework}

Unlike all the existing PCG works, we have proposed a learning-based PCG (LBPCG) framework \cite{b6} that employs machine learning techniques in all the different PCG stages to generate game content from a scratch. The LBPCG framework is motivated by the typical commercial video game development cycles. To address several challenging issues in PCG, the LBPCG explores and exploits knowledge and information gained from game developers and public testers via a data-driven methodology towards minimising interruption to the end-user game-play experience. Relying on generalization of component learning models, the LBPCG tends to direct a game content generator towards generating robust yet appealing content to an arbitrary player. As a result, the LBPCG has several
salient characteristics including avoiding hard coded content evaluation functions, effectively limiting a search space to the relevant content likely to elicit players' specific cognitive and affective experience for adaptivity, and minimising interference
to the end-user game-play experience.

\begin{figure}[t]
\centering
\includegraphics[scale=0.23]{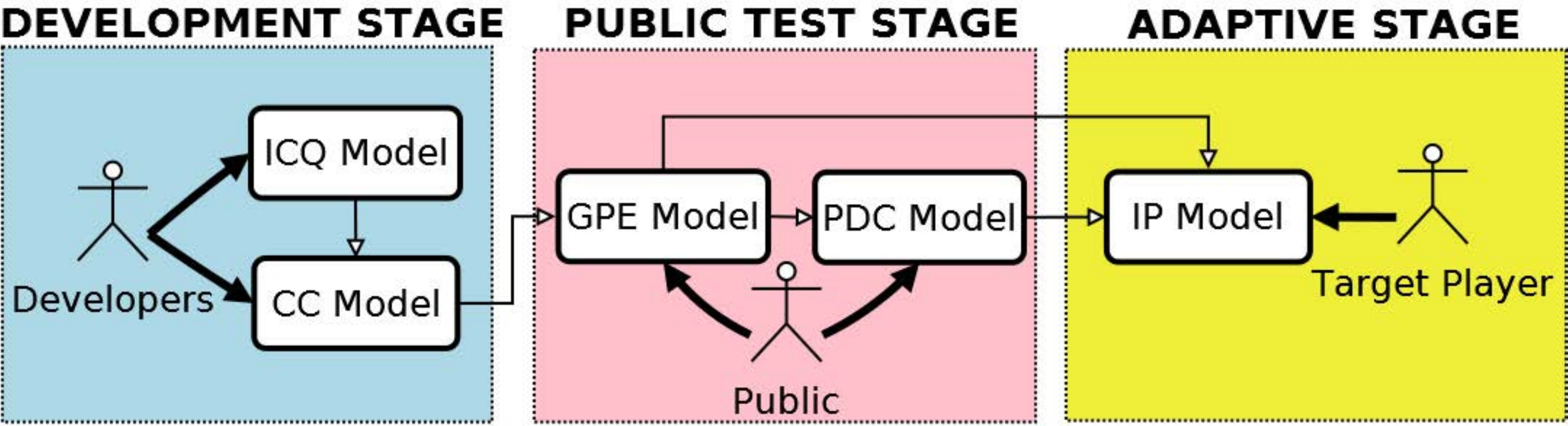}
\label{fig:lbpcg}
\vspace*{-5mm}
\caption{Learning-based procedural content generation (LBPCG) framework.}
\vspace*{-5mm}
\end{figure}

As illustrated in Fig. 1, 
the LBPCG framework consists of three stages: \emph{development} (involving developers), \emph{public test} (involving public testers of different types) and \emph{adaptive} (concerning target players).
In the development stage, we envisage two models, the \emph{Initial Content Quality} (ICQ) and the \emph{Content Categorization} (CC), to encode developers' knowledge. The ICQ model is designed to filter out illegitimate and unacceptable content of poor quality, and then the CC model is used to further partition the legitimate/acceptable content space with pre-defined  content features by developers into meaningful subspaces that is likely to elicit various affective/cognitive experience for players of different types.
In the public test stage, we propose two models, the \emph{Generic Player Experience} (GPE) and the \emph{Play-log Driven Categorization} (PDC). The GPE model is used to capture public players' behaviour and feedback on their experience by playing a number of representative games well selected from different categories defined in the CC model. As public opinions may be different from developers' and their feedback is often subjective and noisy, the GPE model works on finding a ``genuine" consensus on each selected game and assessing the conformity. Thus, the GPE model is expected to find the ``popularity" of any acceptable game and to detect ``outliers" in the cohort of public testers. The PDC model is designed to model cognitive/affective experience based on not only public testers' playing behaviour but also the category of game content that elicits the experience, which forms the key idea in our LBPCG. Thus, the PDC model would predict whether a target player prefers a specific category of games played based on their behaviour after they have played a game in the category. In the adaptive stage, we propose the \emph{Individual Preference} (IP) model to control a content generator
with four models created in the development and the public test stages. The IP model works towards producing their preferred content online for a target player by minimizing interruption to their game-play experience based on all four underpinning models established in developer and public test stages. The IP model addresses several trivial issues: a) automatically detecting
the categorical preference of a target player with the PDC
model and the ``popular" games recommended by the GPE model; b) assuring quality in subsequent generated games within a specific content category with the ICQ, the CC
and the GPE models once the player type is determined; c) automatically
detecting when concept-drift occurs with the PDC
model and tackling it effectively, and d) dealing with crisis situations
autonomously with a system failure avoidance mechanism.

For the LBPCG framework, we develop the enabling techniques for LBPCG by means of  machine learning techniques (c.f. \cite{b6}). A prototype has been established on the first-person shooter, Quake. Over 200 players of different skill levels were involved in our experiments, and experimental results suggest that our LBPCG framework generates quality yet personalised games for players of different types. \cite{b6,b7}.

\subsection{Learning Constructive Primitives for PCG}

While our LBPCG framework is generic to pave a new way for PCG, it needs to be tailored to generate quality game content efficiently and extended to different scenarios such as online game generation and adaptively generating personalised games in real time. Under the LBPCG framework, we have proposed an approach in order to attain the aforementioned goals via learning constructive primitives of video games \cite{b8,b9,b10}. This approach enables online and real-time adaptive video game level generation to be carried out effectively and efficiently from a scratch.

Existing level generators work on a huge content space that contains all the complete procedural levels. As there are an enormous variety of combinations among game elements and structures at a procedural level, such a content space inevitably leads to a greater challenge in managing quality assurance and generating online and adaptive content. In our approach, we decompose a complete procedural level  into a number of short segments as exemplified in Figs 2 and 3. Thus, partitioning a procedural level into fixed-size game segments results in a new content space of lower complexity, which allows us to explore the content space from a different perspective. To this end, a level generation task can be reformulated as finding out a number of quality yet coherent short game segments, named \emph{constrictive primitives} (CPs), to produce a meaning game level as required.
For quality assurance of game levels, there are generally two methodologies in developing such a mechanism in PCG: \emph{deductive} vs. \emph{inductive}. To adopt the deductive methodology, game developers have to understand the content space fully and have skills in formulating/encoding their knowledge into rules or constraints explicitly. In the presence of a huge content space, however, it would be extremely difficult, if not impossible, to understand the entire content space, which might lead to less accurate (even conflicted) rules/constraints used in PCG. On the other hand, an inductive methodology advocated in the LBPCG framework is effective for quality assurance via learning from annotated data. As video game content is observable but less explainable, it is easier for developers to judge the quality of games via visual inspection than to formulate their knowledge into rules formally with sophisticated skills. Thus, a quality evaluation function may be learned from games/levels annotated by developers.

\begin{figure}[t]
\centering
\includegraphics[scale=0.28]{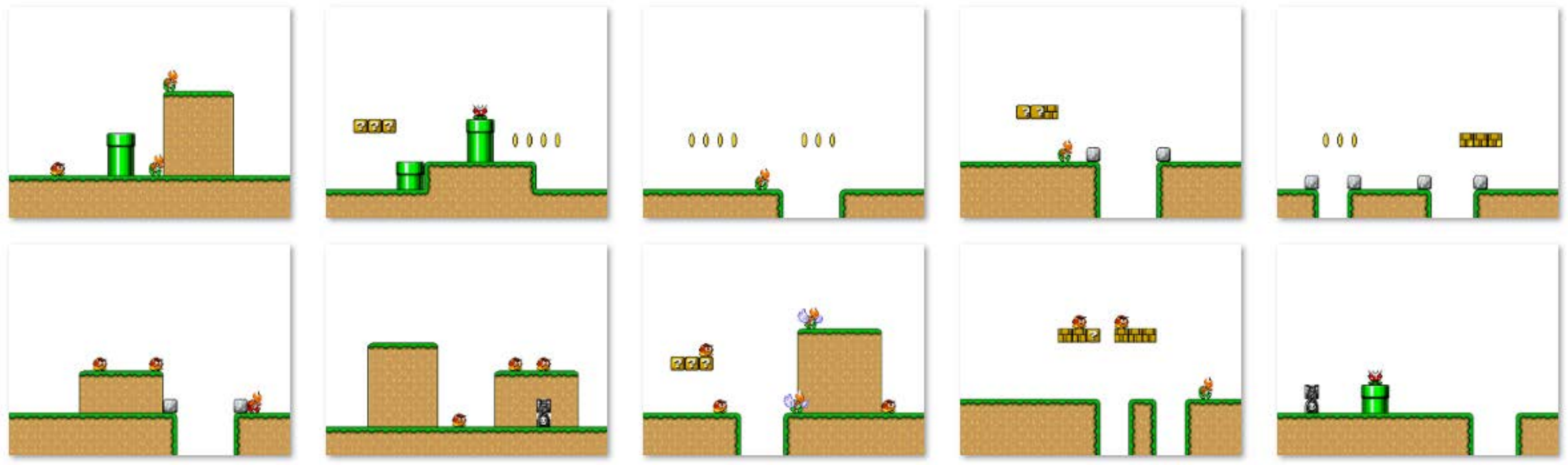}
\label{fig:lbpcg}
\vspace*{-2mm}
\caption{Typical game segments in Super Mario Bros (SMB).}
\vspace*{-1mm}
\end{figure}

\begin{figure}[t]
\centering
\includegraphics[scale=0.12]{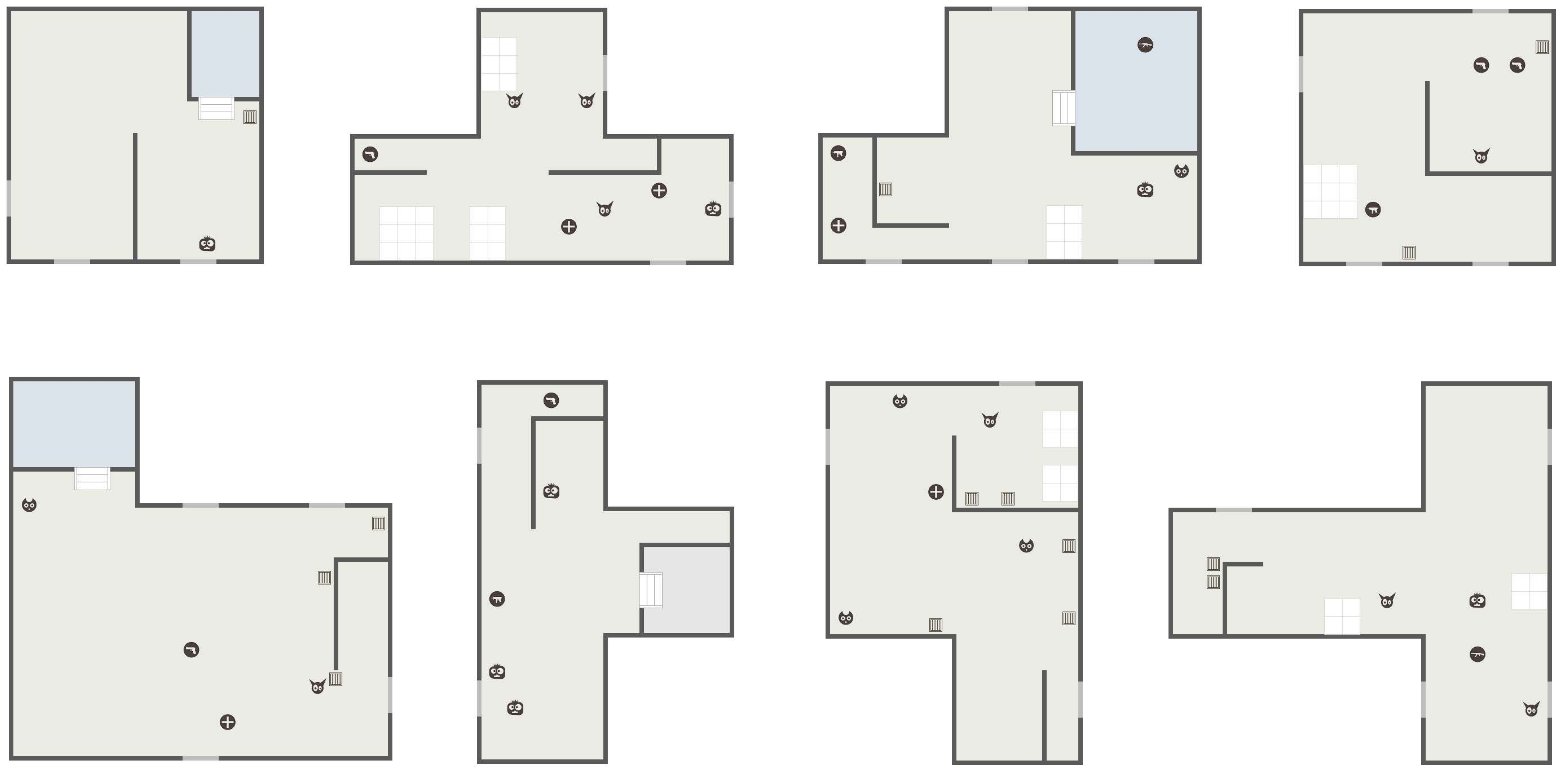}
\label{fig:lbpcg}
\vspace*{-2mm}
\caption{Typical game segments in a first-person shooter game, CUBE 2.}
\vspace*{-2mm}
\end{figure}

\begin{figure}[h]
\centering
\includegraphics[scale=0.29]{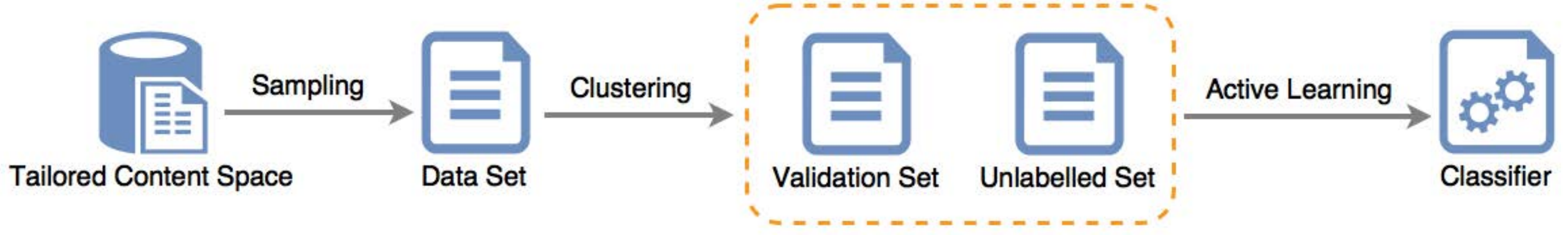}
\label{fig:cp}
\vspace*{-5mm}
\caption{The constructive primitive learning process.}
\vspace*{-0mm}
\end{figure}

We have proposed a generic approach to learning CPs generation \cite{b8,b9,b10}. As illustrated in Fig. 4, our CP learning process consists of three stages: \emph{sampling}, \emph{clustering analysis} and \emph{active learning}.
In the content space, there are a huge number of game segments. To tackle the computationally intractable issue, a proper \emph{sampling} technique is used to generate a much smaller data set of the same properties owned by the original content space. To make a learning-based quality evaluation function, training examples are essential but have to be provided by game developer(s). Although the use of sampling leads to a computationally manageable data set, annotating all the segments in this sampled data set is not only laborious and time-consuming but also may not be necessary if their distribution can be estimated. Thus, we employ \emph{Clustering analysis} for exploring the distribution and the structure underlying the sampled data set. Furthermore, we adopt the \emph{active learning} methodology that enables us to train a classifier with only a small number of most informative game segments annotated by game developer(s). To carry out an effective active learning, we exploit the clustering results to minimize the number of annotated game segments required by active learning since game segments residing in all the different clusters are likely to hold the main properties of the entire tailored content space. Thus, clustering analysis not only facilitates active learning but also reveals non-trivial properties of a content space. Enabling techniques for learning CPs have been developed and the details can be found in
\cite{b8,b9,b10}.

\begin{figure}[t]
\centering
\includegraphics[scale=0.29]{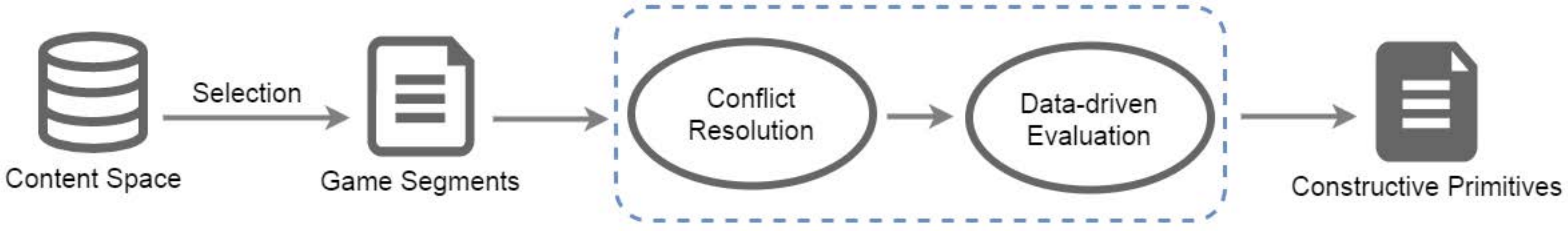}
\label{fig:cp}
\vspace*{-5mm}
\caption{The constructive primitive generation process.}
\vspace*{-5mm}
\end{figure}

\begin{figure}[h]
\centering
\includegraphics[scale=0.29]{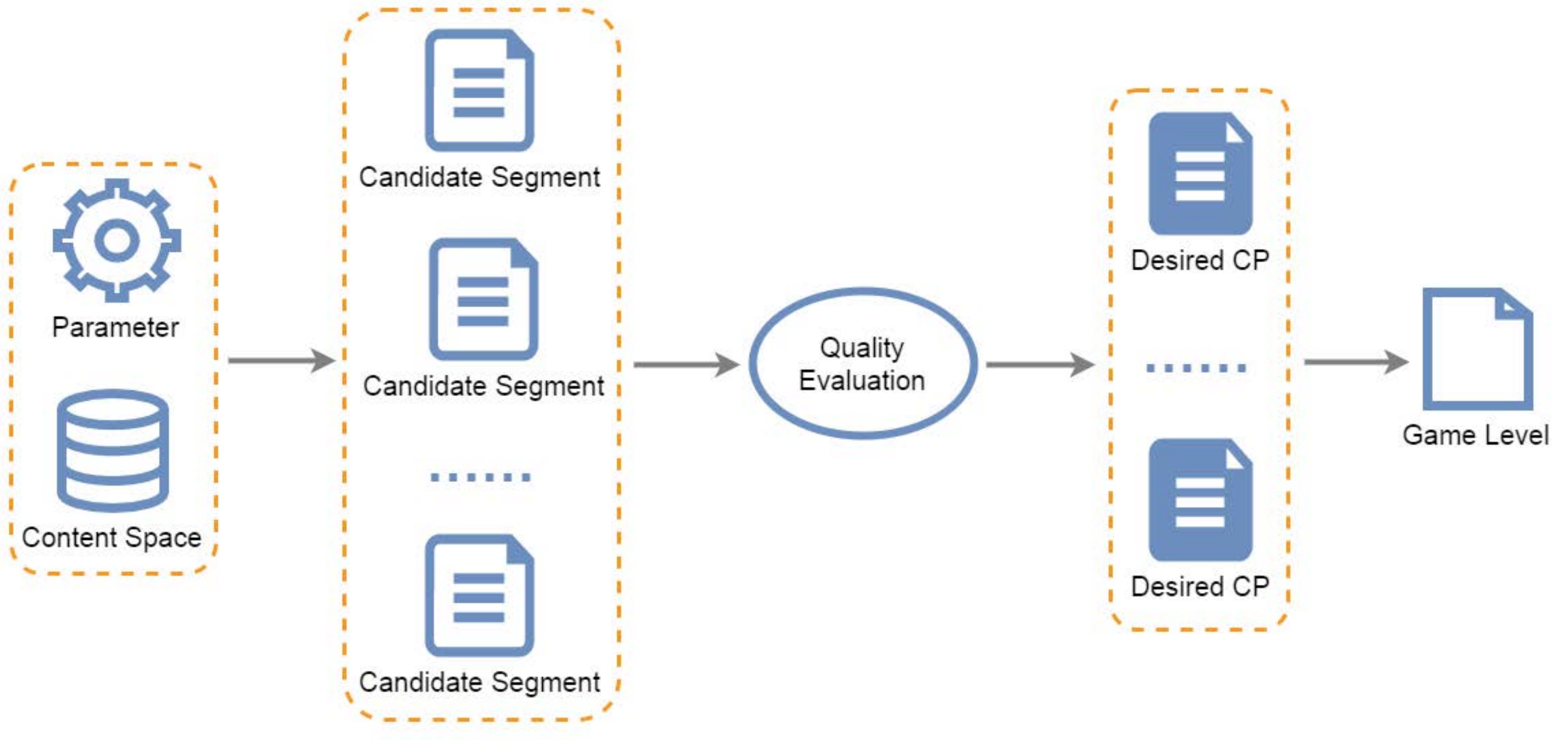}
\label{fig:cp}
\vspace*{-5mm}
\caption{The CP-based online procedural level generation.}
\vspace*{-0mm}
\end{figure}

\begin{figure}[h]
\centering
\includegraphics[scale=0.45]{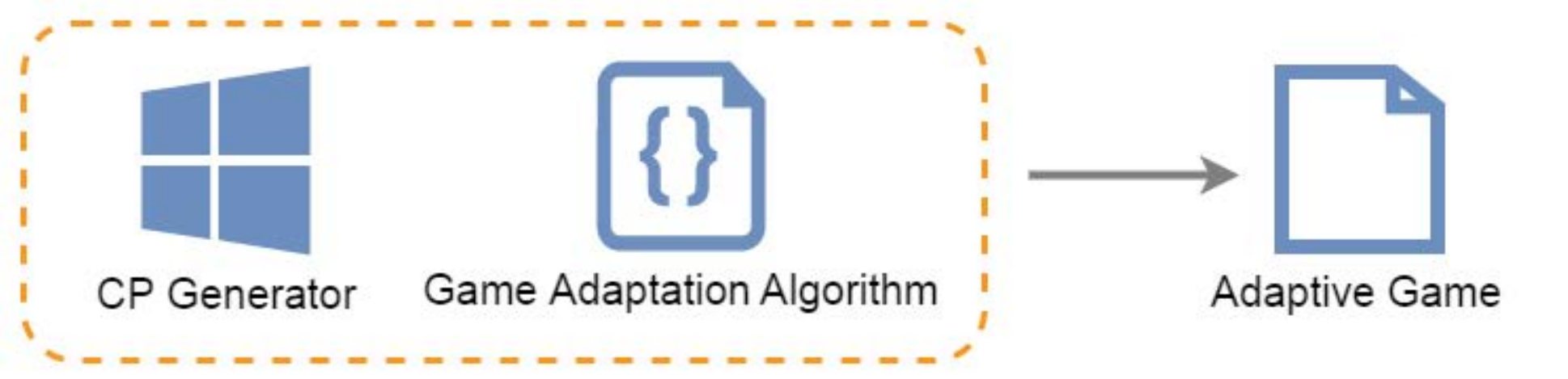}
\label{fig:cp}
\vspace*{-5mm}
\caption{The CP-based adaptive procedural level generation.}
\vspace*{-0mm}
\end{figure}

\begin{figure*}[t]
\centering
\includegraphics[scale=0.85]{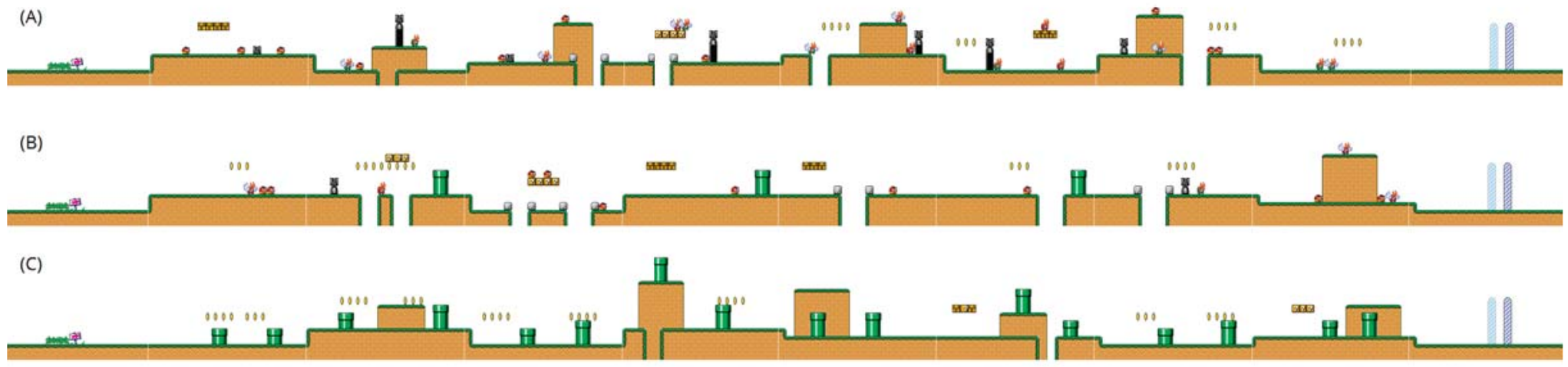}
\label{fig:cp}
\vspace*{-2mm}
\caption{Exemplar SMB procedural levels generated for players of different types. (A) Expert player. (B) Experienced player. (C) Novice player.}
\vspace*{-5mm}
\end{figure*}

Once the active learning is completed, the resultant classifier serving as a quality evaluation function is employed to decide if a short game segment is a CP. As illustrated in Fig. 5, CPs are generated as follows: 1)
all the game segments in the content space are evaluated by a set of easy-to-design conflict resolution rules to eliminate obviously unqualified game element; 2) subsequent machine-learning based quality evaluation function deals with more complicated quality issues; and 3) surviving game segments constitute CPs to facilitate online and adaptive level generation.
This approach exploits the synergy between rule-based and learning-based methods would allow for addressing the quality evaluation issue more efficiently and effectively.

For CP-based online level generation \cite{b8}, a
our CP generator allows game developers to control the low-level geometrical features more easily. Thus, we employ a set of controllable level-generation parameters in order to generate a variety of levels online by setting proper values to relevant content features in CPs.  The controllable parameters depend on game genres. To generate a complete level online, we first specify the desired values for controllable parameters that fix the values of relevant content features and set other irrelevant content features in game segments randomly. Then, game segments with desired properties are examined to generate CPs, and an iterative process is undertaken by merging the CPs with the specified properties. As depicted in Fig. 6, our CP-based online generation algorithm first uses a generate-and-test method to produce CPs for quality improvement, and a complete procedural level is then generated by combining CPs with the specified properties together via setting controllable parameters at a local level. Strategies on merging selected CPs to formulate a complete game differ in game genres (c.f. \cite{b8,b10}). For demonstration, we have developed two CP-based online level generators\footnote{The demo systems and source code are available online: \url{http://staff.cs.manchester.ac.uk/˜kechen/CP-Mario.html} and \url{http://staff.cs.manchester.ac.uk/˜kechen/CP-FPS.html}.} for Super Mario Bros (SMB), a famous 2-D platform game, and CUBE2, a popular first-person shooter game. Detailed results can be found in \cite{b8,b10}.

For CP-based online level generation \cite{b9}, it requires an adaptive strategy to generate a personalised game in real time during game-play, as shown in Fig. 7.
\emph{Dynamic difficulty adjustment} (DDA) is a major methodology used in adaptive content generation where the difficulty of content is dynamically adjusted to match an individual player's skill level and/or to optimize their cognitive/affective experience during game-play \cite{b4}. DDA can be carried out by either a model-free or a model-based methodology. In the model-free methodology, player
model is established for adaptation based on human players’
game-play data and their subjective feedback, while a player model in the model-based methodology is derived from psychological emotion theories. We adopt the model-based methodology and hence take the assumption that maximum game-play enjoyment occurs with the content of a moderate challenge. For the real-time DDA, we have come up with a reinforcement learning algorithm (see \cite{b9,b10} for details) to carry out a model-based DDA algorithm working on CPs for real-time adaptive level generation. This adaptive level generation algorithm generates personalised games in real time based on a player's performance on game segments played in an episode via DDA. For demonstration, we have also developed two CP-based adaptive level generators\footnote{The demo systems and source code are available on the project websites of which URLs are given in footnote 1.} for SMB and CUBE2. Fig. 8 shows exemplar personalised SMB game levels generated in real time by our CP-based DDA algorithm for players of different types during their game-play. More results can be found in \cite{b9,b10}.

\section{Learning-based Serious Education Game}

Serious Educational Game (SEG) \cite{b11} refers to an alternative
learning or supplementary methodology that employs game technologies to primarily
facilitate players’ learning along with gaining positive cognitive
and affective experience during such a learning process.

In SEG development, there are generally two spaces to be taken into account: \emph{knowledge space} regarding learning materials and \emph{content space} regarding games to be used to convey learning materials. The knowledge space is formed to encode learning materials concerning the subject knowledge to be learned by players, while the game content space is formed with playable game elements that convey the knowledge chunks implicitly. Thus, one of the most challenging problems in SEG development is how to deploy the learning materials into game content (formally, how to map the knowledge space to content space) seamlessly, effectively and effortlessly. However, the mapping is well known to be a bottle-neck in SEG development as this has to be handcrafted by game developers closely working with education experts in almost all of existing SEGs, which is costly and labroious. On the other hand, people have different capabilities in knowledge acquisition via gaming SEGs, hence personalised game content matching individuals' capabilities should be generated to ensure that SEGs are indeed an effective yet alternative methodology for learning. To this end, adaptive SEG techniques need to be development. Nevertheless, most of existing SEGs work in a non-adaptive way and hence generate only universal content for different individuals \cite{b12,b13}.  In our study, we have systematically addressed two non-trivial issues by developing a novel SEG framework via effectively mapping knowledge space to game content space exemplified with an SEG,  Chem Dungeon\footnote{The prototype and the source code are available online: \url{https://github.com/mpalmerlee/ChemFight}.}, developed by ourselves \cite{b14} and a learning-based adaptation strategy to personalised SEG generation \cite{b15}.

\begin{figure}[t]
\centering
\includegraphics[scale=0.5]{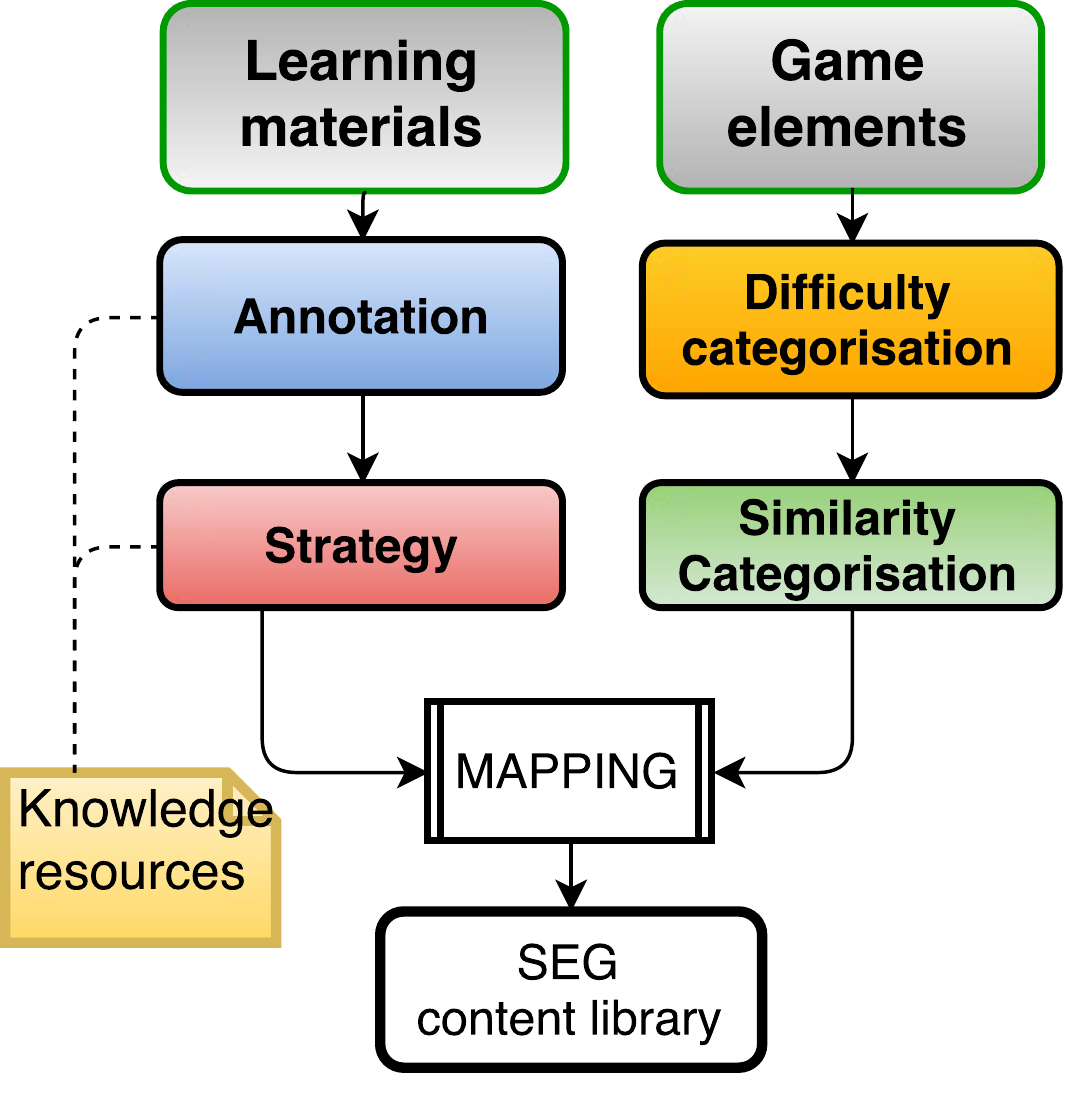}
\label{fig:cp}
\vspace*{-3mm}
\caption{The learning-based SEG development framework.}
\vspace*{-5mm}
\end{figure}

\subsection{Mapping from Knowledge to Content Space}
\label{sect:seg}

Unlike most of the existing SEG approaches, we have proposed an alternative SEG development framework shown in Fig. 9 to address the mapping issue by embedding annotated knowledge chunks into categorized game content/elements seamlessly during SEG development.  As described in Sect. \ref{sect:lbpcg}, PCG techniques may significantly lower the cost of game development. Moreover, the LBPCG framework \cite{b6} suggests that a proper use of the categorized game content may facilitate eliciting positive game-play experience. Thus, our framework would utilise the LBPCG techniques and existing entertainment games in SEG development so that embedding  knowledge chunks into categorised game content/elements would make the mapping easier to accomplish.

As illustrated in Fig. 9, learning materials and game elements are originally in two separate spaces. On one hand, annotation takes place to describe education materials naturally from the meta-data retrievable from reliable resources. Then, we need to establish the strategy for presenting them to players, based on their retrieved properties or use of the corresponding learning resources. On the other hand, categorisation of game content space consists of a couple of steps. It starts with a complexity/difficulty categorisation which groups game content according to the level of challenge. Subsequently, within each of the pre-defined content categories, e.g., a partition of content space in terms of complexity or difficulty levels, clustering analysis is applied to group similar game content for a number of given education materials. Thus, the aspects underlying the descriptive learning materials and game elements can guide a developer to use their logic in formulating the mapping between learning materials and game content. The outcome is an SEG content library comprised of playable games for learning.

Under the SEG development framework depicted in Fig. 9, we have developed an SEG game, Chem Dungeon, to demonstrate its effectiveness. The Chem Dungeon game is a single-player game against a non-player character enemies, which allows one to learn chemistry knowledge by collecting compound-forming atom objects and entering the exit gate within a specified time limit.
To develop this game, we use the library of education materials and the basic rule (pairing atoms to create a compound) as the knowledge space and an existing rogue-like game to form the content space. All the resources regarding the knowledge and the game content spaces are available publicly. As shown in Fig. 10, the game field has pathways and walls that form a maze with intersections and cul-de-sac. An exit gate, initially closed, is hidden at the bottom-right of the maze. Actors in the game consist of an avatar and some opponents, each with a spawn point. The avatar carries an atom within its shield where the corresponding information is located nearby its spawn point.
There is a button to open the periodic table and a \texttt{Help} button to pause the game and show mission objectives. Meanwhile, information regarding a compound-forming result or an atom properties is at the top-centre of the game arena. The right side of the game contains lives, experience progress, ammunition, the remaining time and the total correct compounds collected.
Inside the maze, bullets, atom objects and life potions are collectable for the avatar. Each bullet collected adds some ammunition for the avatar. A bottle of potion can restore the avatar's life to full. The score gained by a player reflects how much knowledge they have learned.

\begin{figure}[t]
\centering
\includegraphics[scale=0.42]{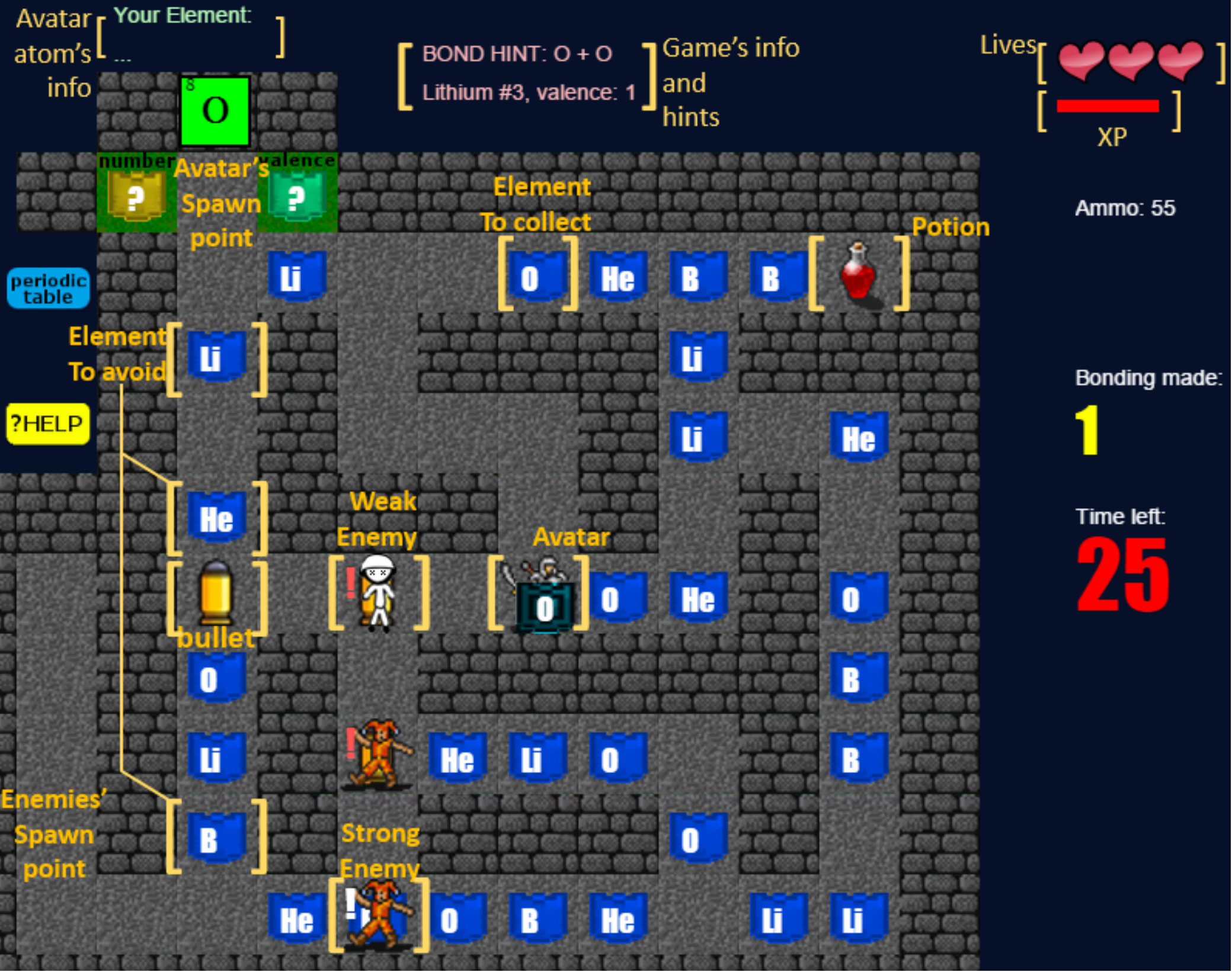}
\label{fig:cp}
\vspace*{-5mm}
\caption{The Chem Dungeon platform layout.}
\vspace*{-5mm}
\end{figure}

To evaluate this game, we have conducted a user study of 50 players via a survey containing the SEG allows players to play the game and report their game-play experience, e.g., ``fun". Examinations are used to assess the learning performance of the players, where each question in the examination represents a piece of learning material. Thus, pre- and post-game examinations produced binary values indicating prior knowledge and recalling results, respectively. The difference in scores between pre and post-game examinations produced three types of learning performances: unchanged, improvement and decay. The experimental results suggest that the Chem Dungeon game considered as successful from the players’ perspective regarding their learning and enjoyment (c.f. \cite{b14}).

\subsection{Strategy for Adaptive Serious Education Game}

To generate the personalised games, we have proposed an adaptation strategy by considering an individual's learning/game-play capabilities as well as their experience based on our own SEG framework described in Sect. \ref{sect:seg}.

\begin{figure}[t]
\centering
\includegraphics[scale=0.47]{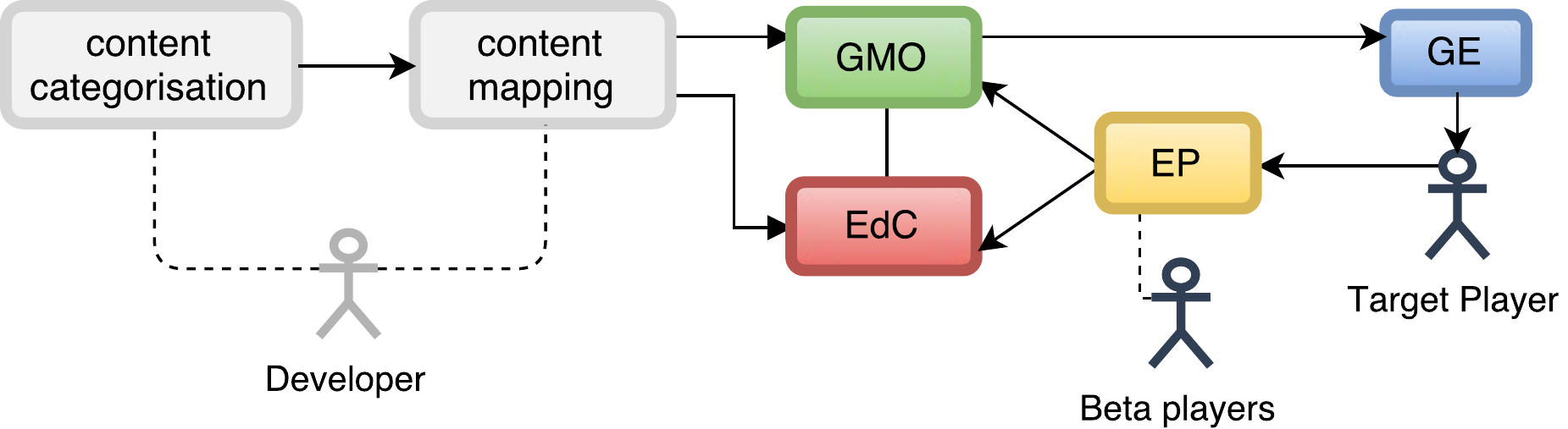}
\label{fig:cp}
\vspace*{-5mm}
\caption{The strategy for adaptive serious education games.}
\vspace*{-5mm}
\end{figure}

As illustrated in Fig. 11 (see \cite{b15} for details), our adaptation strategy organises SEG content by \emph{content categorisation} in knowledge and content spaces, followed by \emph{content mapping} to ensure that a knowledge chunk uniquely corresponds to selected game content of all difficulty levels. Then, the adaptation focuses on the \emph{game mission optimisation} (GMO) and \emph{experience-driven content} (EdC). The GMO is used to personalise the sequence of game stages that optimises the achievements of the player with respect to their learning and enjoyment, while the EdC is designed for effective yet
efficient game content search based on the player’s experiences.
The GMO works in the top level of the content space of the SEG; i.e., education materials, and the EdC searches the best game content. The adaptivity is continuously progressing, given that the player's \emph{game-play experience} (GE) are assessed by the \emph{experience prediction} (EP) in a non-intrusive way via learning from gaming data such as play-log and features of game content, and passes them to GMO and EdC. Iteratively, the GMO receives a pair of inputs from EP's prediction and the played game stage(s); it then updates its knowledge about the pair and optimises the future generation of the learning tasks. Subsequently, EdC considers the predicted experience received from the EP and the similarity of the specific game content corresponding to a learning task.

Based on the above adaptation strategy, we have conducted simulations on our SEG described in Sect. \ref{sect:seg}, \textit{Chem Dungeon}, to demonstrate the usefulness via reinforcement-learning based agents simulating players of different types. The simulation results indicate that the proposed adaptation strategy is effective in improving the learning performance of players of different types (see \cite{b15} for detailed results).

\section{Learning-based Fast Skill Capture}

Skill is an important component of video game-play activities. Not only does it contribute to the outcome, but also the relationship between skill and the difficulty of the activities affects the affective/cognitive experience of players'. More specifically, players in a video game often have the most fun when
their skill is matched to proper game content equally, rather than dominating novices or being dominated by highly accomplished players \cite{b16}. However, measuring a player's skill is time consuming and hence cannot be done until a player completes a substantial of various games \cite{b17}. On the other hand, many aspects in video game development, such as adaptive elements of game levels, characteristics of agents' behaviour, and player matching in multi-player games, require a manner that can measure a player’s skill level rapidly. To address this challenge, we have conducted a systematical study with machine learning and exploratory data analysis for fast skill capture from players' input and behaviour recorded in their play-log \cite{b18,b19,b20}.

\begin{figure}[t]
\centering
\includegraphics[scale=0.35]{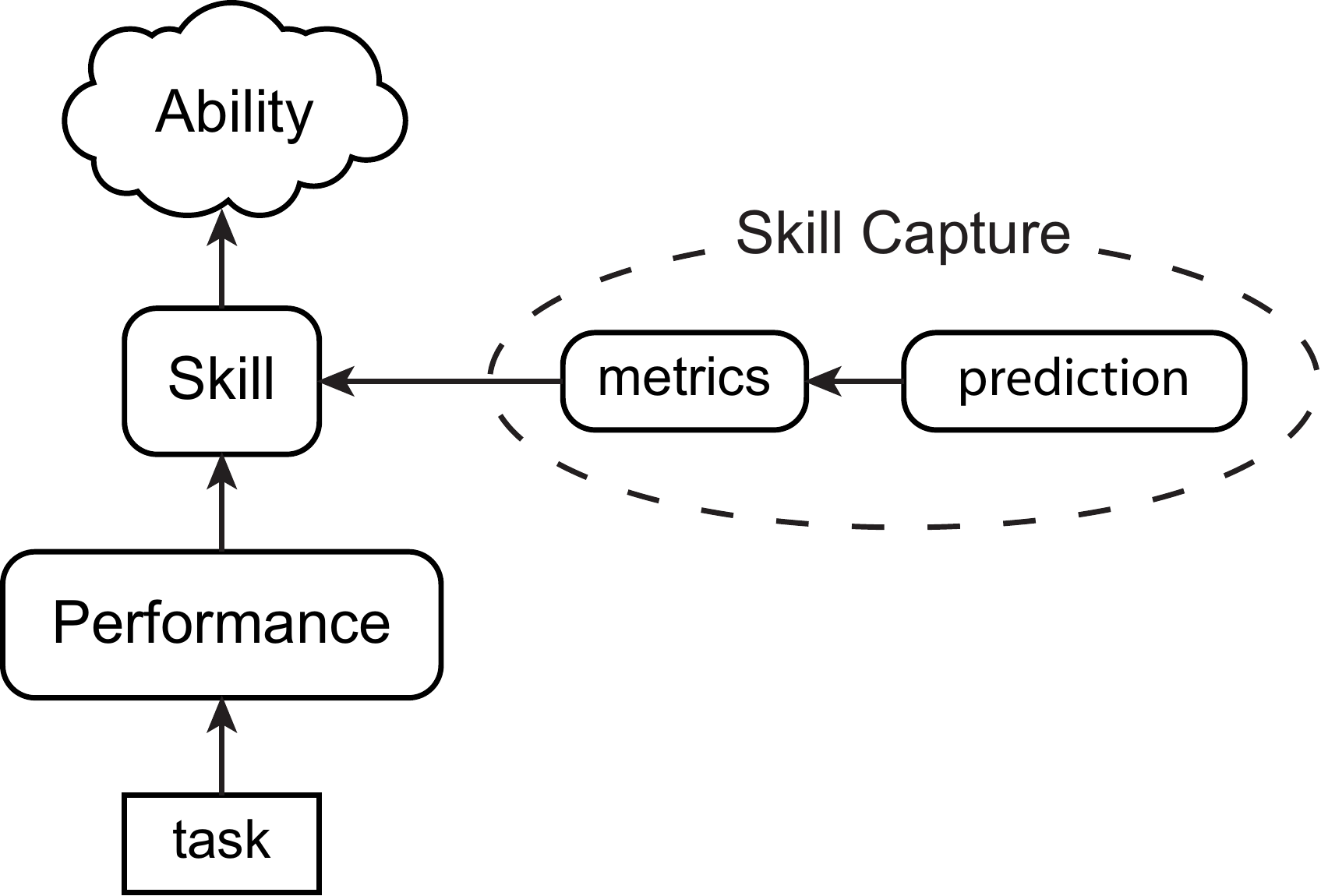}
\label{fig:cp}
\vspace*{-2mm}
\caption{The schematic of our fast skill capture process.}
\vspace*{-5mm}
\end{figure}

For fast skill capture, we define skill as a property of a player in terms
of their average performance. we have observed that a player's input, e.g., mouse and key presses, is consistent over several games although as their performance may depend on several factors including their opponents. Therefore, we assume
that a skilled player would interact with the controls differently
to a novice. Instead of relying on performance as a
measure for each player, we consider using their input and hypothesize
that it could better predict skill from fewer games.
Towards this goal, we have performed a systematic study
based on a first-person shooter. Game logs
were automatically recorded during the study, storing input
events, some game events and a few common measurements of
performance. In order to understand these measurements, we
present a thorough analysis of those aspects and the features extracted
from the input events with exploratory data analysis techniques . By means of machine learning techniques, e.g., random forest, we then predict the player’s skill with reasonable accuracy from a short period of data regarding a player's input and behaviour. As depicted in Fig. 12, our skill capture model\footnote{The prototype and other resources regarding this research are available online:
\url{http://www.cs.man.ac.uk/~buckled8/shortterm.html}.} readily predicts a player’s skill within a single game.

For a thorough evaluation, we have used keyboard and mouse data to predict
the player’s skill in different ways and achieved some favourable results as follows. Primarily, we find that it is more feasible to predict a player’s skill within 30 second of game data with the player’s input than it is using the player’s performance for that game. In addition to predicting categories of players such as skilled player vs. novice, we can also predict continuous skill measures such as a player’s average score and so on (see \cite{b19,b20} for detailed results).

\section{Object-based Learnable Agent}

General video-game AI (GVGAI) \cite{b21} is regarding the
development of general AI algorithms that enable autonomous learnable agents to
play a wide range of video-games with minimal tailoring
to specific games. Such agents are required to replace hand-coded in-game AI and to serve as a development tool or a proxy for human play-testers in video game development. Furthermore, GVGAI has wider implications for the field of AI, as techniques working well on video-games can often be applied to related real-world
problems. For GVGAI, a substantial progress has been made via \emph{reinforcement learning} (RL), e.g., RL-based agents achieve a level comparable to the performance of professional human games testers on Atari games \cite{b22}. Nevertheless, the RL-based agent \cite{b22} works in an end-to-end manner from pixels to actions, which is significantly distinct from what human players do in a manner of using prior knowledge about the world to make efficient decision making \cite{b23}. Abundant evidence suggests that human players work on objects in a video game via perceptual organisation of coherent pixels rather than treating all the pixels in a frame independently. To this end, we have developed object-based reinforcement learning techniques \cite{b24} to create ``human-like" learnable agents.

\begin{figure}[t]
\centering
\includegraphics[scale=0.25]{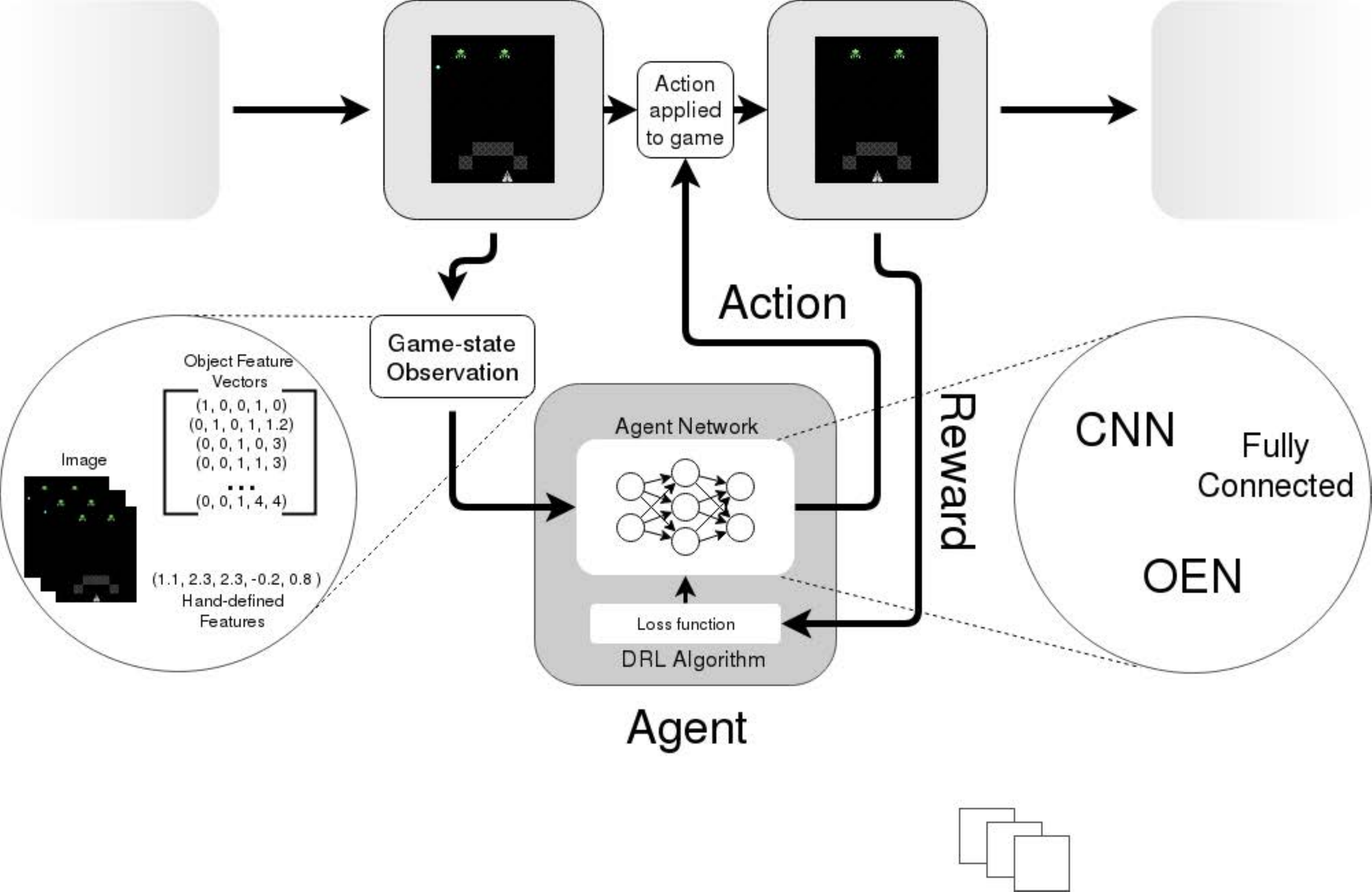}
\label{fig:cp}
\vspace*{-5mm}
\caption{The object-based learnable agent via reinforcement learning.}
\vspace*{-0mm}
\end{figure}

\begin{figure}[t]
\centering
\includegraphics[scale=0.36]{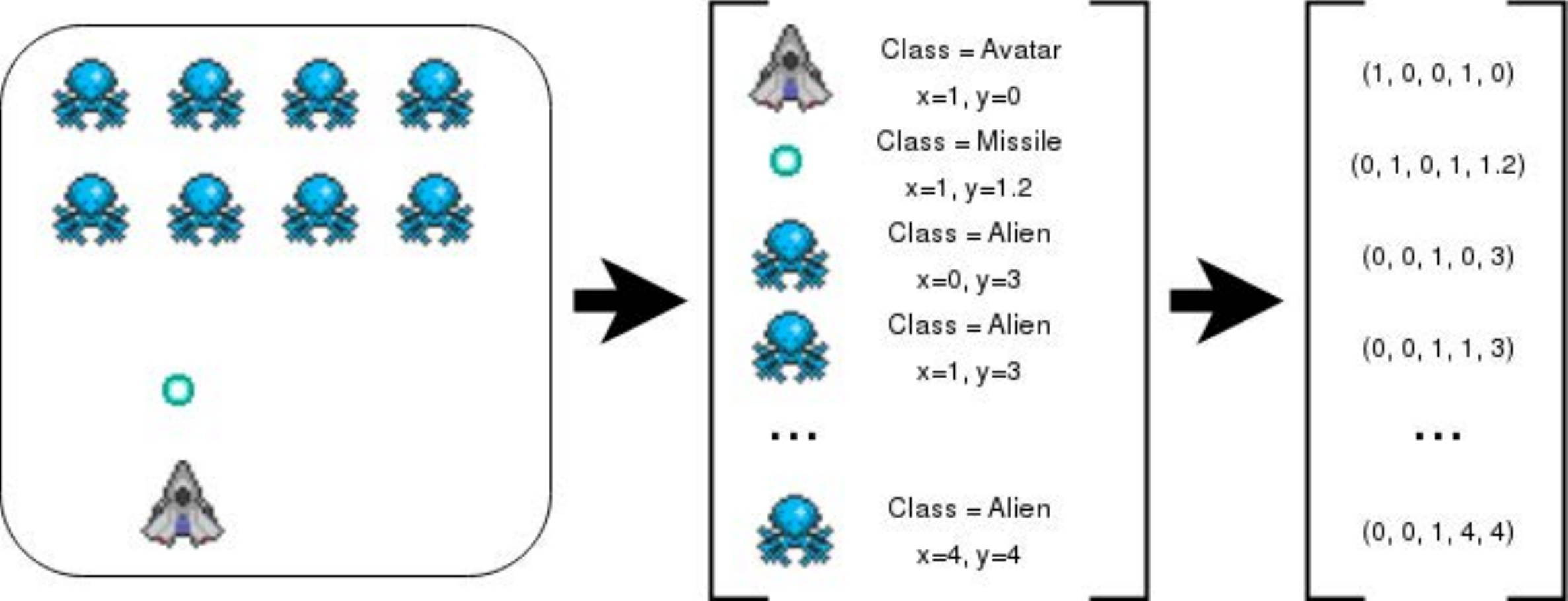}
\label{fig:cp}
\vspace*{-2mm}
\caption{The set of objects in a game state and their internal representation.}
\vspace*{-5mm}
\end{figure}

We have developed an object-based learnable agent\footnote{The prototype and source code are available online: \url{https://github.com/EndingCredits/Object-Based-RL}.} based on deep RL techniques. As illustrated in Fig. 13, our agent works on a set of objects in a game state,  one frame in a game video clip, to find out a ``proper" (optimal) action in this game state to gain a ``high" reward (score) with a deep neural network trained on historical playing data via RL.
One of the biggest challenges in creating an object-based learnable agent is that
there are a variable number of objects in different game states, e.g., a typical game state in Atari games where the number of aliens is various in differen game states, as exemplified in Fig. 14. If we use deep RL techniques to learning an object-based agent, we have to address this challenge as a deep neural network can cope with only the input of a fixed length. Thus,
how to represent a variable number of object in different game states in a video is a critical issue for object-based RL. In our work, we have come up with a novel approach to address this issue in GVG-AI. As a result, we adopt a specific type of neural network architecture, set networks, to develop a novel
object embedding network (OEN) as shown in Fig. 15. The OEN can not only
take a list of object-feature vectors of arbitrary lengths as input
to produce just a single, yet unified, fixed-length representation
of all the objects within the current game state, but also be
trained on a given task simultaneously. Therefore, our OEN-based learnable agent works with objects, rather than raw pixel data, can help scale up deep learning to more complicated tasks in a similar fashion to human information processing.

\begin{figure}[t]
\centering
\includegraphics[scale=0.4]{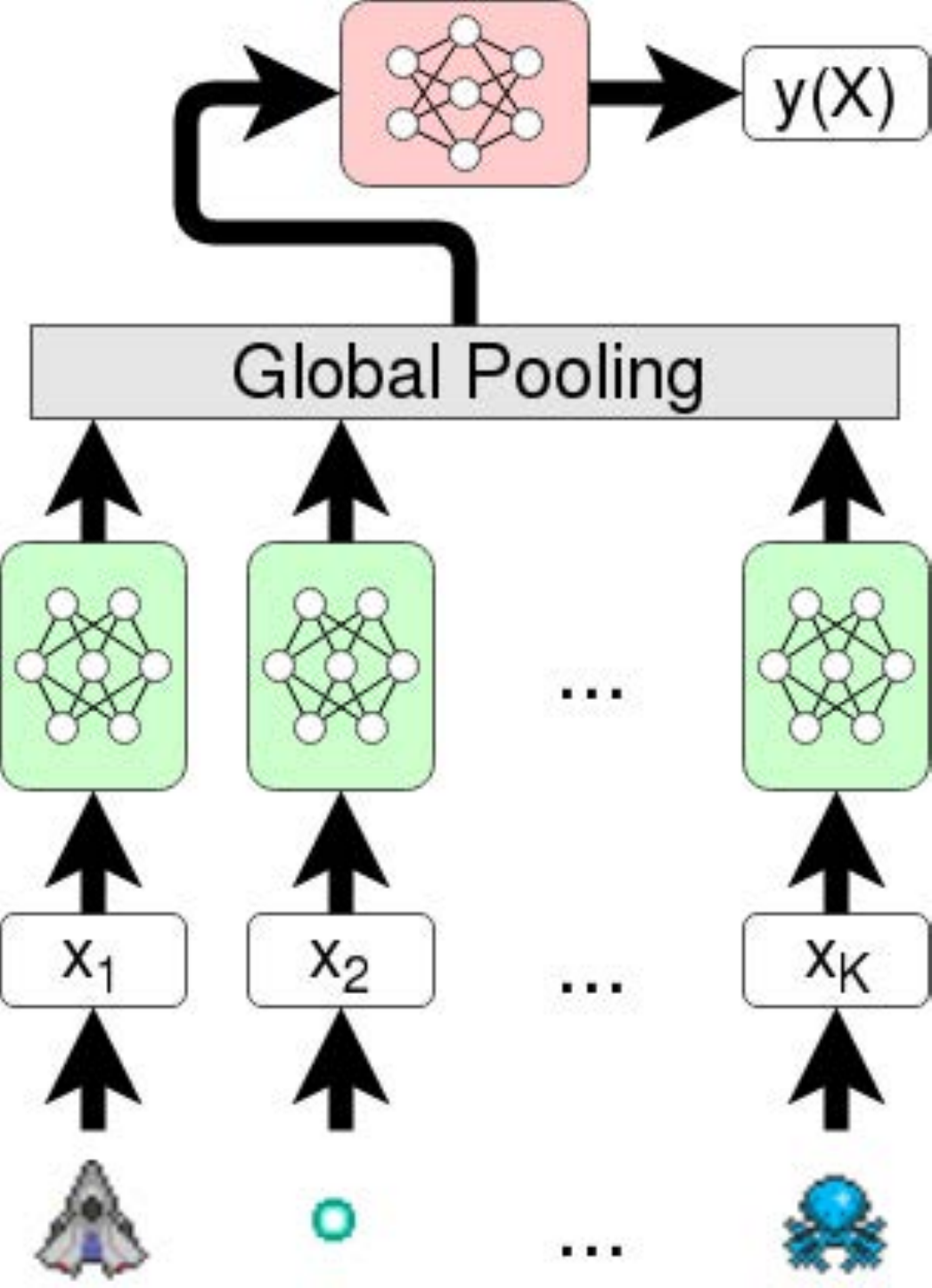}
\label{fig:cp}
\vspace*{-2mm}
\caption{The object-embedding network architecture for object-based RL.}
\vspace*{-5mm}
\end{figure}

We have conducted experiments on five selected games used in the GVGAI
competition \cite{b21} in comparison to those RL agents training on raw video data under the same conditions. From experimental results, we do observe that there are notable difference in performance between pixel-level and object-based representations. It is evident from the experimental results that our object-based learnable agent is capable of autonomous learning in all five games and leads to better generalisation performance or robustness immune to noise and the background variation in raw video data (see \cite{b24} for detailed results).

\section{Outlook}

As described in the previous sections, substantial progress in several areas of video game development have been made by means of machine learning in the MLP@UoM lab. Below we briefly discuss the future research directions arising from our works.

Our LBPCG framework paves an alternative way for PCG to allow for generating video games from scratch. However, our framework is dependent on the quality of each of its constituent models to have good performance. Therefore, developing more effective enabling techniques and other safeguard measures needs to be undertaken in the future. Another challenge for our LBPCG is how to acquire the quality training data required by all the component models. In our current works, the data labelling and annotation have to be done by developers and public testers. However, this could still be a burden in developing large-scale and complex video games. Fortunately, this issue could be addressed by autonomous learnable agents who can mimic real players' behaviour and user modelling via the state-of-the art machine learning.

To develop SEGs effectively towards the ultimate goals, our proposed framework by mapping knowledge to game content relies on the sophisticated LBPCG techniques as well as automatic acquisition of knowledge chunks . Although we have successfully developed an SEG under this framework, effective enabling techniques are still missing and many technical issues remain unsolved. Furthermore, our adaptive SEG strategy aims assessing the player’s learning and enjoyment without interrupting the game session for generating personalised games. However, this strategy working on several assumptions has yet to test with real players. Given the fact that players' feedback is subjective hence could be noisy, non-trivial issues for this strategy are expected in real applications and have to be addressed in future study.

Fast skill capture is a relatively new area in video game development and there are many avenues of research left open to explore. While our learning-based fast skill capture approach works well for a first-person shoot, it has to be assessed in other game genres. Also, our skill prediction model is heavily based on the players' input and their feedback but has yet to explore other gaming data sufficiently and consider those factors external to the game at all. Again, how to acquire training data efficiently is still open to solution, Furthermore, our approach works only on single-player games while such an approach is highly demanded for player-matching in multi-player games. All the above issues have to be tackled in future research.

Our object-based learnable agents aims working in a human-player style based on RL. However, a human player is capable of making good use of prior knowledge and geometric/semantic relationship between different objects in a game state. To this end, our object-based model has yet to explore/exploit the aforementioned aspects. Novel enabling techniques need to be developed to carry out human-like functions. Moreover, learnable agents who can mimic human players' behaviour are highly demanded in video game development such as LBPCG, learning-based SEGs and fast skill capture. To the best of our knowledge, all the existing learnable agents aims the optimal performance during game-play and never take into account simulating the behaviour of human players of different types ranging from novice to expert. Thus, creating the ``human-like" learnable agents with appropriate machine learning techniques would be a very important research area in future.

\section*{Acknowledgment}

The work presented in this paper was collaborated with those MLP@UoM members who worked in learning-based video game development. The author is especially grateful to his former PhD students, J. Roberts, P. Shi, H. Rosyid, D. Buckley and W. Woof, for their contributions.




\begin{thebibliography}{00}
\bibitem{b1} L. Galway, D. Charles, and M. Black, ``Machine learning in digital games: A survey,'' AI Review, vol. 29, pp. 123--161, 2008.
\bibitem{b2} G. Yannakakis and J. Togelius, Artificial Intelligence and Games. Springer, 2018.
\bibitem{b3}
J. Togelius \emph{et al}, ``Search-based
procedural content generation: A taxonomy and survey,''
IEEE Trans. Comput. Intell. AI Game., vol.~3,  pp. 172--186, 2011.
\bibitem{b4}
G. Yannakakis and J.~Togelius, ``Experience-driven procedural content
  generation,'' IEEE Trans. Affect. Comput., vol.~2,  pp. 147--161, 2011.
\bibitem{b5}
A. Summerville \emph{et al}, ``Procedural content generation via machine learning (PCGML),'' IEEE Trans. Game., vol.~10,  pp. 256--270, 2018.
\bibitem{b6} J. Roberts and K. Chen, ``Learning-based procedural content generation," IEEE Trans. Comput. Intell. AI Game., vol. 7, pp. 88--101, 2015.
\bibitem{b7} J. Roberts, Learning-based Procedural Content Generation. PhD Thesis,
School of Computer Science, The University of Manchester, 2014.
\bibitem{b8}
P. Shi and K. Chen, ``Online level generation in Super Mario Bros via learning constructive primitives," In Proc. IEEE CIG'16, 2016.
\bibitem{b9}
P. Shi and K. Chen, ``Learning constructive primitives for real-time dynamic difficulty adjustment in Super Mario Bros,'' IEEE Trans. Game., vol.~10,  pp. 155--169, 2018.
\bibitem{b10} P. Shi, Learning Constructive Primitives for Procedural Content Generation. PhD Thesis, School of Computer Science, The University of Manchester, 2019.
\bibitem{b11} D. Michael and S. Chen, Serious Games: Games that Educate, Train, and Inform, Thomson Course Technology, 2006.
\bibitem{b12} D. Ismailovic \emph{et al}, ``Adaptive serious game development," in: Proc. IEEE Int. Workshop Game. Software Eng., 2012.
\bibitem{b13} A. Hussaan, K. Sehaba, and A. Mille, ``Tailoring serious games with adaptive pedagogical scenarios: A serious game for persons with cognitive
disabilities," in: Proc. ICALT, 2011.
\bibitem{b14}
H. Rosyid, M. Palmerlee, and K. Chen, ``Deploying learning materials to game content for serious education game development: A case study,'' Entertainment Computation, vol.~26,  pp. 1--9, 2018.
\bibitem{b15} H. Rosyid, Adaptive Serious Educational Game with Machine Learning. PhD Thesis, School of Computer Science, The University of Manchester, 2018.
\bibitem{b16}   M. Csíkszentmihályi, Flow: The Psychology of Optimal Experience.
 Harper \& Row, 1990.
\bibitem{b17} P. Dangauthier \emph{et al}, “TrueSkill
through time: Revisiting the history of chess,” In Advances in Neural
Information Processing Systems 20, 2008.
\bibitem{b18} D. Buckley, K. Chen, and J. Knowles, ``Predicting skill from player input in a first-person shooter," In Proc. IEEE CIG'13, 2013.
\bibitem{b19} D. Buckley, K. Chen, and J. Knowles, ``Rapid skill capture in a first-person shooter," IEEE Trans. Comput. Intell. AI Game., vol. 9, pp. 63--75, 2017.
\bibitem{b20} D. Buckley, Skill Capture in a First-Person Shooter. PhD Thesis, School of Computer Science, The University of Manchester, 2016.
\bibitem{b21} D. Perez \emph{et al}, ``General video game AI: A multi-track tramework for evaluating agents, games and content generation algorithms,'' IEEE Trans. Game., vol.~11, 2019. (to appear)
\bibitem{b22} V. Mnih \emph{et al}, “Human-level control through deep reinforcement learning,” Nature, vol. 518, no. 7540, pp. 529–533, 2015.
\bibitem{b23} R. Dubey \emph{et al}, “Investigating human priors for playing video games,” In Proc. ICML'18, 2018.
\bibitem{b24} W. Woof and K. Chen, ``Learning to play general video games via an
object embedding network," In Proc. IEEE CIG'18, 2018.


%
\end{thebibliography}
\end{document}